\documentclass{article}

\usepackage{arxiv}

\usepackage[utf8]{inputenc} % allow utf-8 input
\usepackage[T1]{fontenc}    % use 8-bit T1 fonts
\usepackage{hyperref}       % hyperlinks
\usepackage{url}            % simple URL typesetting
\usepackage{booktabs}       % professional-quality tables
\usepackage{amsfonts}       % blackboard math symbols
\usepackage{nicefrac}       % compact symbols for 1/2, etc.
\usepackage{microtype}      % microtypography
\usepackage{lipsum}
\usepackage{fancyhdr}       % header
\usepackage{graphicx}       % graphics
\graphicspath{{media/}}     % organize your images and other figures under media/ folder

\title{Language, Environment, and Robotic Navigation
}

\author{
  Johnathan E. Avery \\
  Vivum AI \\
  \texttt{jack@vivum.ai} 
}

\begin{document}
\maketitle

\begin{abstract}
This paper explores the integration of linguistic inputs within robotic navigation systems, drawing upon the symbol interdependency hypothesis to bridge the divide between symbolic and embodied cognition. It examines previous work incorporating language and semantics into Neural Network (NN) and Simultaneous Localization and Mapping (SLAM) approaches, highlighting how these integrations have advanced the field. By contrasting abstract symbol manipulation with sensory-motor grounding, we propose a unified framework where language functions both as an abstract communicative system and as a grounded representation of perceptual experiences. Our review of cognitive models of distributional semantics and their application to autonomous agents underscores the transformative potential of language-integrated systems. 
\end{abstract}

% keywords can be removed

\section{Introduction}

There is much research exploring the development of autonomous agents capable of navigating their environments through both direct interaction and linguistic inputs \cite{zghair_one_2021,zhang_survey_2022, crespo_semantic_2020}. This paper aims to survey current robotic navigation methodologies, with a special emphasis on those integrating language and semantic information. The primary focus is on Neural Network (NN) models and Simultaneous Localization and Mapping (SLAM), highlighting their distinct advantages and potential synergies with linguistic data. Finally, we discuss cognitive models of distributional semantics that extract map information from text, and look at ways to incorporate approaches in cognitive science to autonomous agent applications.

Neural Network approaches leverage deep reinforcement learning algorithms, allowing robots to dynamically learn from and adapt to their surroundings by processing extensive sensor data to inform navigation decisions. This method demonstrates exceptional utility in dynamic environments requiring real-time responsiveness. In contrast, SLAM enables robots to concurrently generate a map of their environment and ascertain their location within it, facilitating navigation in unfamiliar or evolving areas absent pre-existing maps. Additionally, path planning algorithms, such as A* \cite{liu_comparative_2011} and Dijkstra's algorithm \cite{soltani_path_2002}, offer methods for calculating the most efficient paths between points, while behavior-based systems merge simple navigational behaviors, such as obstacle avoidance and goal seeking, to traverse complex environments.

Our discussion examines current work that integrates language and semantics into NN and SLAM frameworks. We aim to incorporate cognitive models of distributional semantics to enable autonomous agents to extract map information from textual data, bridging the gap between cognitive science principles and autonomous agent applications. We expect that this cross-disciplinary approach will enrich the robot's navigational capabilities and open avenues for more intuitive human-robot interactions.

\subsection{Incorporating Language into Robotic Navigation}

Simultaneous Localization and Mapping (SLAM) is an approach in robotics that enables autonomous agents to navigate their environments . Traditional SLAM focuses on constructing a map of an environment while simultaneously determining the robot's location within that map, using sensor data from cameras, LiDAR, or other devices. This process primarily concerns itself with geometric and spatial data \cite{levine_learning_2023}, creating a structural representation of the environment without necessarily assigning meaning to the objects within it \cite{seymour_maast_2021}. As such, traditional SLAM provides the robot with the capability to navigate but not necessarily to interact with its surroundings in a contextually meaningful way.

Semantic SLAM (S-SLAM) extends this concept by integrating semantic understanding into the mapping process \cite{bowman_probabilistic_2017}. It not only identifies the locations and shapes of objects but also assigns labels or categories to them, turning a purely geometric map into a semantic one. This integration relies heavily on machine learning models, particularly those trained on vast datasets with labeled examples of objects and their attributes. However, one limitation of S-SLAM is its dependency on human-annotated data for training these models, which can be resource-intensive and may not capture the full complexity and variability of real-world environments. Additionally, while S-SLAM enables robots to recognize and classify objects, it often still treats language and semantic information as separate from the core navigational and operational algorithms, primarily using it to enhance object recognition and scene understanding \cite{drouilly_semantic_2015}.

Neural network approaches to robot navigation have evolved to leverage deep learning for both scene interpretation and deciding on actions. These systems can process complex sensor inputs to directly inform navigational decisions, learning optimal paths and behaviors from experience rather than relying solely on pre-defined maps or models. While powerful, these approaches have traditionally focused on optimizing for specific tasks or environments, often without a deep integration of semantic understanding into the representational space of the agent.

Neural network approaches incorporating language seek to bridge the gap between linguistic input and an agent's sensorimotor experience. By training autonomous agents in simulated environments to locate specific objects described by various adjectives, researchers establish demonstrate that a link between linguistic descriptors and the agent's perceptual experiences can resolve the symbol grounding problem \cite{hermann_grounded_2017, ghaffari_grounding_2023}. Additionally, language has been used as a command interface, where instructions are provided to the agent, enabling complex tasks such as manipulating objects with a robotic arm \cite{lynch_interactive_2022, bing_meta-reinforcement_2022} or navigating through simulated environments \cite{fu_language_2019}. These approaches not only enhance the agent's ability to perform tasks based on verbal commands, allowing for more contextually aware robotic navigation.

The proposition of integrating distributional semantic approaches into robot navigation represents a nuanced shift from these methodologies. Unlike semantic SLAM, which may rely heavily on predefined labels and human annotations, distributional semantics offers a way to derive the meaning and relationships of entities directly from their occurrence and co-occurrence in data—akin to how language models learn the semantics of words from the contexts in which they appear. By embedding these principles into neural network models for navigation, the goal is to enable robots to infer functional and relational properties of objects and spaces in a more autonomous, data-driven manner. This would allow robots to navigate their environment not just in terms of physical layout or labeled categories, but as a network of meaningfully related concepts and entities.

Such an approach diverges from using language merely as a command interface for robots; instead, it integrates linguistic and semantic knowledge into the robot's decision-making processes. This integration facilitates the robot's ability to infer purposes, potential actions, and relationships without explicit human instruction for each scenario. The goal is to create robotic agents that can navigate and operate in complex, dynamic environments with a high degree of complex behavior, leveraging the messiness and richness of linguistic data to inform their perceptions and actions in the physical world.

\section{SLAM}

SLAM involves two primary tasks: localization (determining the robot's position within the map) and mapping (constructing a map of the environment). These tasks are performed simultaneously based on sensor data from cameras, LiDAR (Light Detection and Ranging), or other sensing technologies. The SLAM process typically involves the following steps:

\begin{enumerate}
    \item \textbf{Data Collection}: The robot collects data from its environment using sensors.
    \item \textbf{Feature Extraction}: It extracts features from the sensor data, which could be specific points or landmarks in the environment.
    \item \textbf{Data Association}: The robot identifies which features in the current data correspond to previously observed features.
    \item \textbf{State Estimation}: Using the associated data, the robot updates its estimate of its own position and the positions of features in the environment. This often involves probabilistic models to account for the uncertainty in sensor data and feature identification.
    \item \textbf{Map Update}: The map is updated with the new data, refining the representation of the environment.
    \item \textbf{Loop Closure Detection}: The system detects if the robot has returned to a previously visited area, which can significantly reduce cumulative mapping errors.
\end{enumerate}

SLAM algorithms can be implemented using various approaches, including Extended Kalman Filters (EKF), Particle Filters (Monte Carlo methods), and Graph-Based methods. Each of these approaches deals differently with the uncertainty inherent in sensor data and motion.

\subsection{Semantic SLAM}

The semantic information in Semantic SLAM systems can be derived from a combination of sources, and the exact methodology often depends on the complexity of the system, the specific application, and the available resources. Here are the primary sources of semantic information:

\begin{enumerate}
    \item \textbf{Pre-defined Knowledge Graphs and Ontologies}: These are structured frameworks that provide detailed information about various objects, their attributes, and the relationships between them, facilitating behaviors in complex environments by providing a rich semantic context \cite{walter_learning_2013}. For instance, if a robot recognizes an object as a chair, the knowledge graph can inform it that a chair is meant for sitting, and it's typically found in certain types of rooms.

    \item \textbf{Machine Learning Models}: Deep learning and other machine learning models are often trained on large datasets to recognize and classify objects within an environment. These models can identify objects (like tables, chairs, doors) and even more abstract concepts (like navigable space) based on visual, auditory, and tactile data. The semantic labels assigned by these models are then integrated into the SLAM map, providing a layer of understanding about what the objects are and potentially their function or purpose.

    \item \textbf{Human Input and Interaction}: In some systems, semantic information is enriched or initiated through direct human input or interaction \cite{lynch_interactive_2022}. This can involve a user labeling parts of the environment or teaching the robot about different objects and their uses. With sufficient data, the robot can generalize from these interactions to facilitate navigation of the environment.

    \item \textbf{Online Databases and APIs}: Some systems might access online databases or use APIs to fetch real-time information about objects or scenes. For example, a robot might recognize a specific object but then query an online database for information regarding its uses, dimensions, or other relevant semantic information.

    \item \textbf{Sensor Data Fusion}: Semantic information can also be derived from the fusion of different types of sensor data, like visual, LiDAR, or radar data \cite{gervet_navigating_2022}. This approach allows the system to exploit the strengths of each sensor type, leading to more accurate and detailed semantic maps.
\end{enumerate}

In practice, advanced Semantic SLAM systems might use a combination of these sources to create robust and informative maps that not only navigate the physical space but also understand the function and context of the environment and its contents.

\subsubsection{Ontological Approaches}
\label{sec:ontology}

Ontologies, as formal tools for describing objects, properties, and relationships within a knowledge domain, offer a structured way to model the world with which a robotic system might interact. Ontologies serve as explicit, formal specifications of shared conceptualizations or logical theories that articulate what a set vocabulary intends to convey \cite{studer_knowledge_1998, guarino_formal_1998}. These structured systems enable the categorization and hierarchical organization of knowledge, making use of classes to represent concepts at various levels, relations to denote associations between these concepts, and formal axioms to ensure consistency and coherence among the relationships. By employing such a structured approach, ontologies facilitate the representation of the environment in a manner that is both comprehensive and understandable to machines, typically utilizing languages like first-order logic for formal specification. This level of organization is crucial in robotics for building models of the environment where relevant concepts are clearly defined and interrelated, allowing for efficient navigation, interaction, and task execution within that environment.

\subsubsection{Semantic Maps}

Semantic maps in the context of robot navigation integrate complex semantic concepts of the environment, such as the identification of objects, their functionalities, and the categorization of different spaces based on their uses and human-like perceptions \cite{hemachandra_learning_2015}. These maps go beyond traditional metric and topological representations by incorporating high-level semantic concepts that facilitate interaction with humans and improve the robot's service tasks. Semantic maps encapsulate valuable environmental information through associations between objects and their features, elevating the map's utility for autonomous navigation and task execution \cite{crespo_semantic_2020}. This approach relies on a structured understanding of space, where properties such as object presence, door locations, room shapes, sizes, and appearances are annotated and associated with sensory data to create a nuanced, multi-layered representation of the environment \cite{kostavelis_learning_2013, han_semantic_2021}. In practice, semantic maps have facilitated robot navigation of indoor \cite{alenzi_semantic_2022} and outdoor \cite{karimi_semantic_2021} environments.

\subsubsection{S-SLAM vs Our Approach}

Our approach, while acknowledging the significance of semantic maps in enhancing robot-environment interaction, aims to diverge towards a more fluid understanding of semantic relationships through distributional semantics. This paradigm shift emphasizes deriving meaning and relationships from the agent's interaction with both the physical and linguistic environments, transcending the more static associations found in traditional semantic maps. Here, ‘semantic’ extends to encompass the dynamic and context-dependent interplay between concepts, grounded in the robot's experiences and interactions within its environment. This method seeks to embed a richer, more adaptable layer of semantic understanding into the robot's navigational and operational framework, aspiring for a more human-like comprehension of environmental semantics where meanings are not just assigned but discovered and evolved over time.

Our distributional approach to robot navigation and understanding diverges significantly from the rigid, axiomatic structures of ontologies \cite{jamieson_instance_2018}. Rather than pre-defining and categorizing the world into a fixed hierarchy of concepts and relationships, the distributional approach focuses on learning from the agent's continuous interaction with both its physical and linguistic environments. This approach is underpinned by the symbol interdependency hypothesis \cite{louwerse_symbol_2011}, which suggests that the meanings of symbols (or words) emerge from their use in varied contexts, rather than from their assignment to pre-defined categories. Distributional semantics, therefore, emphasizes the derivation of meaning from the statistical properties of symbols as they appear and co-occur in natural language, allowing an agent to form an understanding of the world that is fluid and adaptable. This adaptability is key in dynamic and complex environments where rigid categorizations may fail to capture the nuanced interplay of objects and actions. By prioritizing experience and interaction over predefined structures, we expect that the distributional approach will enable robots to navigate their surroundings in a way that is more reflective of human cognitive processes, where meanings and relationships are constantly negotiated and redefined through engagement with the world.

\section{Neural Network Approaches}

Neural network (NN) models represent a shift from traditional SLAM algorithms in robot navigation, emphasizing learning from data over probabilistic modeling and geometric analysis. Unlike SLAM's reliance on explicit models for localization and mapping, NNs excel in deriving complex patterns and relationships directly from data. This distinction underscores NNs' versatility in adapting to diverse environments without being constrained by the need to model uncertainty or geometry explicitly as in SLAM.

NN approaches introduce a dynamic and flexible framework for robot navigation. For example, Convolutional Neural Networks (CNNs) enhance feature extraction capabilities, enabling more nuanced scene interpretation, and Recurrent Neural Networks (RNNs) facilitate sequence data processing, improving localization and mapping efforts. This integration of NNs into the SLAM process signifies the potential for a symbiotic relationship where NNs augment SLAM's geometric precision with adaptive, data-driven insights.

The integration of language into NN robotic navigation is distinct from SLAM, particularly in how they incorporate language for navigational tasks. NN models utilize language in various innovative ways, from serving as an oracle in reinforcement learning scenarios where linguistic cues guide reward mechanisms \cite{guo_improve_2023}, to processing complex commands for direct action \cite{fu_language_2019}. Integrating language into NN approaches enable agents not only navigate but also interpret and act within their surroundings in a contextually rich manner \cite{hermann_grounded_2017}.

\subsection{Language as Command Interface}

Integrating language into robot or simulated agent behavior growing field. Commonly, language is implemented as a command interface as a means to guide and direct robot behavior \cite{nair_learning_2022}. This approach seeks to bridge the communication gap between humans and robots, making interactions more intuitive and accessible. Instead of relying on pre-programmed instructions or complex programming languages, robots can interpret verbal or written instructions, translating the complexities of human language into actionable tasks.

Numerous approaches have successfully demonstrated that incorporating language as the command interface for autonomous agents improves task completion in unfamiliar environments in imitation learning \cite{zhou_language-conditioned_2024}, meta-reinforcement learning \cite{bing_meta-reinforcement_2022}, and inverse-reinforcement learning \cite{fu_language_2019}.

Goyal et al. introduce PixL2R, a model that innovatively uses natural language descriptions to map visual inputs directly to rewards in reinforcement learning environments, thereby facilitating policy learning without predefined reward structures \cite{goyal_pixl2r_2021}. This approach enhances the agent's exploration efficiency by generating intermediate rewards based on the alignment between executed trajectories and linguistic task descriptions, a method proving to significantly increase sample efficiency across various tasks. PixL2R demonstrates the potential of natural language to not only direct but also accelerate the learning process, challenging traditional, hand-designed reward systems with its adaptability and effectiveness in both sparse and dense reward scenarios. Moreover findings by \cite{fu_language_2019} suggest that using language-conditioned rewards offer a more flexible and transferable approach to task specification in reinforcement learning, outperforming direct language-conditioned policies in terms of generalization across new scenarios.

Language as a command interface facilitates robot behavior and navigation by translating human verbal or written instructions into actionable tasks, allowing robots to understand and execute complex commands in their operating environments. This approach leverages natural language processing to bridge the gap between human communication and robotic actions, making interactions more intuitive and expanding the robots' capabilities to navigate and perform tasks with minimal human intervention.

\subsection{From Command Interface to Representational Space}

Building on the foundation of language as a command interface, we propose evolving this concept towards language as a representational space within robotic systems. This expanded paradigm aims not only for robots to understand and execute commands but also to enable them to produce meaningful language in response to their environments. This approach envisions a more integrated and reciprocal relationship between language and sensory experiences, allowing robots to not only act upon language commands but also to describe, reason, and learn from their environmental interactions through language.

In this context, language may serve as a medium for bidirectional communication, enhancing the interaction between humans and robots beyond simple command execution. By treating language as a communication interface, the focus shifts towards facilitating a two-way dialogue where robots can provide feedback, ask for clarification, and share observations, making the interaction more dynamic and collaborative. This is a significant departure from the narrow interaction paradigm of language as a mere command interface, which traditionally directs communication from the human user to the agent in a unidirectional manner.

Furthermore, conceptualizing language as a representational space within robotic systems allows for a more profound integration of linguistic and modal experiences. In this integrated framework, language processing and sensory processing are not isolated; instead, they inform and enhance each other. Such a system enables robots to not only process and respond to linguistic and sensory inputs concurrently but also to use language to fill gaps in their knowledge and to interpolate between linguistic and sensory experiences in a meaningful way. This symbol interdependency between language and modal experience enables robots to navigate and interact with their environment in a more nuanced and human-like manner, paving the way for more sophisticated and autonomous robotic agents capable of complex reasoning and learning.

\section{Maps, Language, and Cognition}

\subsection{Learning Semantic Maps From Natural Language}

Cognitive maps highlight the usefulness of language, emphasizing the redundancy between our use of words in contexts and their physical environmental counterparts. 

The incorporation of language into robotic navigation systems has primarily focused on addressing the symbol grounding problem \cite{hermann_grounded_2017, chan_zipfian_2022, ghaffari_grounding_2023}, or on implementing language as a command interface for agents. Alternatively, research has shown that language serves as a substrate to communicate features of a map effectively.  On a local level, one can convey information about a location with a single statement (e.g., "Your cup is on the left"). On a larger scale, language patterns robustly support map features, such as city distributions. Research using Latent Semantic Analysis (LSA), an early model of distributional semantics based on word co-occurrences, demonstrated that word representations for cities accurately reproduced a map of the United States \cite{louwerse_language_2009}. A follow-up study used the "Lord of the Rings" trilogy as a text corpus \cite{louwerse_representing_2012}, showing that LSA could reliably reconstruct a map of Middle Earth purely from the text. Furthermore, the maps derived from LSA closely matched those created by human participants tasked with the same activity, suggesting a similarity in cognitive map construction processes between humans and the distributional semantic algorithm. Cognitive maps highlight the usefulness of language, emphasizing the redundancy between our use of words in contexts and their physical environmental counterparts.

Follow-up research pinpointed the specific statistical distributions facilitating the operation of LSA and other distributional semantic algorithms \cite{avery_reconstructing_2021}. Even when the map was not fully articulated by the corpus, the map was reliably reconstructed, showcasing the effectiveness of language in encoding relationships even with a high sparsity of references. Follow up demonstrations with transformers showed that the embedding space alone could usefully reconstruct maps, and that the downstream attention and feed-forward computations enabled the model to produce accurate and relevant utterances \cite{avery_representation_2023}.

\subsection{Symbol Interdependency}

The distinction between amodal, symbolic cognition, and grounded, embodied cognition is a cornerstone concept in cognitive science and artificial intelligence \cite{de_vega_symbols_2012}. Here's an articulation of the various sides of the debate.

\subsubsection{Symbolic cognition entails the manipulation of amodal symbols}

Symbolic cognition revolves around the idea that human thought processes can be emulated through the manipulation of symbols that stand in for objects, concepts, and ideas in the real world. These symbols are "amodal" in that they do not inherently possess sensory attributes such as sight, sound, or touch. Instead, they function through an abstract representation, allowing for complex thought processes, reasoning, and problem-solving in a manner that is decoupled from the physical characteristics of the environment.

\subsubsection{Symbolic cognition works because the relationships between concepts is preserved, despite the concepts themselves not being grounded}
  
The power of symbolic cognition lies in its ability to preserve the web of relationships between concepts. This preservation enables the manipulation and examination of these relationships independently of the physical or sensory characteristics of the objects or concepts they represent. It's this abstraction that allows for high-level reasoning and abstract thought, key aspects of human intelligence.

\subsubsection{Grounded cognition entails the relationships of physical or modal environments}

Grounded cognition contrasts with symbolic cognition by anchoring concepts and their relationships within the sensory and physical experiences of the world. Here, knowledge and thought are directly tied to sensory inputs and motor outputs, suggesting that understanding stems from embodied experiences. Grounded cognition posits that concepts are understood in terms of their relationships to physical or modal environments, which means cognition is heavily influenced by an individual's sensory and motor interactions with their surroundings.

\subsection{Reconciling Symbolic and Embodied Cognition: Insights and Challenges}

Symbolic cognition offers a framework for understanding and emulating human intelligence through the manipulation of abstract symbols and their interrelationships, without the need for these symbols to be directly grounded in sensory experiences. This contrasts with grounded cognition, which emphasizes the importance of sensory and motor experiences in shaping understanding and thought processes. Both approaches offer valuable insights into the nature of cognition, suggesting that human thought may encompass elements of both symbolic manipulation and sensory-motor grounding.

From the perspective of embodied cognition, the major drawbacks to symbolic cognition center on its disconnection from sensory and motor experiences, which are fundamental to cognition. Embodied cognition critiques argue that without symbol grounding, where words and concepts are tied to bodily actions and sensory experiences in the environment, symbolic cognition remains an endless loop of defining symbols with other symbols, akin to a "symbolic merry-go-round" \cite{harnad_symbol_1990}. This critique posits that comprehension cannot be achieved through mere manipulation of amodal symbols, as it overlooks the crucial role of perceptual experiences and motor actions in understanding language.

Importantly, experimental evidence supports the embodied cognition view by demonstrating that comprehension involves the reenactment of sensory and motor experiences. For example, neurophysiological studies (e.g., \cite{pulvermuller_brain_2005}) have shown that processing action words activates brain areas associated with the corresponding motor actions, indicating a direct link between linguistic stimuli and sensorimotor processes. Comprehension of language not only involves linguistic regularities but also relies heavily on the activation of perceptual systems.

Embodied cognition argues that the abstract and arbitrary nature of linguistic symbols, as treated in symbolic cognition, cannot form the basis of meaning on their own. Instead, meaning arises from the interaction of these linguistic symbols with embodied representations, suggesting that language comprehension and conceptual processing are intrinsically linked to sensory and motor experiences. This perspective challenges the efficacy of purely symbolic models and emphasizes the importance of grounding linguistic symbols in physical and modal experiences to achieve genuine comprehension and meaningful interaction with the environment.

\subsection{Bridging Symbolic and Embodied Cognition for Robotic Navigation}

The symbol interdependency hypothesis \cite{louwerse_symbol_2011} proposes a synthesis where symbolic and embodied cognition are not seen as mutually exclusive but rather as complementary and mutually reinforcing. This perspective acknowledges that while symbolic cognition operates through the interdependencies among amodal linguistic symbols, embodied cognition grounds these symbols in perceptual and sensory experiences. Symbol interdependency suggests that language comprehension involves both symbolic processes, through the relationships among abstract symbols, and embodied processes, through the references these symbols make to tangible, perceptual representations.

This hypothesis highlights that language naturally evolves to encode relationships that exist in the world, including those grounded in physical experiences, thereby serving as a communicative shortcut that encapsulates both symbolic and perceptual information. This dual nature of language, encoding both amodal symbolic patterns and modal, perceptual information, offers a powerful framework for understanding how language can encode information about the world, such as action dynamics and object characteristics, in a way that is readily accessible and easily communicated.

For integrating language into robotic navigation, the symbol interdependency hypothesis offers a foundational basis for developing systems that can interpret and use language in a way that reflects both its symbolic and embodied aspects \cite{luo_transformer-based_2024}. By recognizing that language can encode perceptual information, robotic systems can be designed to leverage linguistic data to enhance spatial understanding and navigation tasks \cite{huang_visual_2023}. Such systems could utilize linguistic input to supplement sensor data, enriching the robot's model of the environment with information that is implicitly encoded in language, such as spatial relations and object attributes. We expect that this approach will not only enhance the robot's ability to navigate through its environment but also facilitates more natural and intuitive human-robot interaction, bridging the gap between high-level linguistic communication and low-level sensory-motor tasks.

\section{Integrating Symbol Interdependency and Robotic Navigation}

In our exploration of the symbol interdependency hypothesis, we anticipate implementing and investigating its principles through the lens of autonomous agents navigating a linguistically enriched and modal environment. At the heart of our approach lies the integration of abstract linguistic inputs with the tangible sensory data of simulated 3D landscapes, aimed at embodying the symbiotic essence of symbolic and embodied cognition. By designing agents with the capability to process both visual and linguistic stimuli, underscored by convergence zones for seamless integration, we apply our understanding of human cognitive processing within a structured model. We foresee that engaging with linguistic environments will not only bolster the agents' modal navigational abilities but also foster a bidirectional reinforcement between linguistic learning and sensory experience. Moreover, we posit that immersion in the modal environment should suffice for agents to articulate precise linguistic descriptions independently, grounded in their direct and descriptive environmental learning. We aim to illuminate the practical applicability of the symbol interdependency hypothesis, bridging theoretical cognition with the tangible implementations of linguistic input and sensory experience in the domain of autonomous navigation systems.

\subsection{Agent and Environment}

Our investigation of the symbol interdependency hypothesis is aimed at the interplay between linguistic and modal environments within the framework of autonomous agents. This exploration delineates two environmental inputs that an agent encounters: the linguistic environment and the and the modal environment. The linguistic environment encompasses language inputs provided to the agent and subsequent language the agent generates. The modal environment is defined as a simulated 3D space, likely a series of landmarks in order to align with ongoing research into linguistic cognitive maps (rather than room-to-room navigation as is conventional in robot navigation literature).

At the core of our architectural design for such agents is the processing systems of both visual (or scene-related) stimuli and linguistic data. This dual-input system is supported by several key modules, which, though not specified in detail within this paper, are essential for processing both visual scenes and linguistic information. 

Importantly, the agent architecture will include convergence zones—dedicated areas within the agent's framework that integrate both linguistic and visual inputs. We expect that these zones are instrumental in facilitating a seamless interaction between the scene and language processing modules. We acknowledge the complexities and limitations inherent in simulating human cognition. Recognizing that human cognition is not entirely modular, we note that certain brain regions have stable roles in processing specific information types. 

Our agents will employ an encoder-decoder setup where the encoder provides context for the decoder, allowing it to process previously produced tokens and generate the most appropriate subsequent token. This setup reflects findings which illuminate the insufficiency of sparse rewards in reinforcement learning tasks for achieving language grounding—the crucial connection between linguistic symbols and their corresponding visual stimuli \cite{hermann_grounded_2017}.

Through this structured exploration, we aim to contribute to the understanding of how autonomous agents can effectively navigate and interact within their environments by leveraging the symbiotic relationship between linguistic and modal inputs. This approach not only seeks to enhance the agents' navigational abilities but also underscores the potential for more nuanced and contextually rich human-robot interactions, bridging the gap between high-level linguistic communication and sensory-motor functionalities.

\subsection{Testing Symbol Interdependency}

Given the symbol interdependency hypothesis, we anticipate three outcomes in our agents. First, we would expect that experience with the linguistic environment should facilitate the agent's ability to navigate the modal environment. Second, we would expect that learning in both the linguistic and modal domains should be mutually reinforcing. Third, we would expect that learning in the modal environment should be sufficient to enable an agent to produce accurate linguistic statements without direct linguistic training.

\begin{enumerate}
    \item \textbf{Learning Language from a Corpus}: We would expect agents that are trained on a corpus that describes an environment to learn to navigate the environment more efficiently than an agent with no prior training on corpus statistics. 
    \item \textbf{Learning Language/Distributional Information from Environmental Interaction}: This involves agents generating something akin to an internal monologue based on their visual experiences, assuming a pre-existing mapping between visual stimuli and word labels. The sequence of visual labels in the agent's monologue is expected to reflect the physical arrangement of objects in its environment. We would expect an agent that learns semantic representations from their experiences should learn to navigate its environment more effectively than the same agent that does not learn semantic representations based on interaction with its environment.
    \item \textbf{Communicating Representations Learned from Environmental Interaction}: Under the symbol interdependency hypothesis, we envision overlapping sets of symbols and modal experiences, where some symbols are directly grounded in modal interactions, yet some modal experiences lack direct linguistic expression. We would expect that an agent exposed to both linguistic and modal environments is expected to produce more accurate utterances than agents exposed to a singular (even if more complete) environment, benefiting from the integration of information across domains.
\end{enumerate}

These points aim to articulate the expected cascade of information between modal and symbolic processing, testing the validity of the symbol interdependency hypothesis in bridging linguistic and sensory experiences within cognitive models.

\section{Conclusion}

We aim to advance the integration of linguistic inputs into the navigational capabilities of autonomous agents, guided by the symbol interdependency hypothesis. Our approach delineates the distinct but interconnected roles of linguistic and modal environments in shaping cognitive models for robotic navigation.

Looking ahead, we intend to implement and test the outlined architecture, focusing on the impact of integrating linguistic and visual inputs within the agents. Key to our future direction is the examination of how agents can leverage linguistic data to enhance spatial understanding and decision-making processes, thereby enriching their interaction with the environment.

The significance of this work extends beyond the immediate advancements in robotic navigation; it contributes to the broader discourse on cognitive models, offering insights into the complex interplay between language and sensory experiences. By grounding our investigation in the symbol interdependency hypothesis, we underscore the potential for linguistic data to transform the way autonomous systems navigate and interpret their environments. This research paves the way for more intuitive and natural human-robot interactions, marking a significant step towards the realization of truly autonomous agents capable of navigating the world with the same nuance and understanding as humans.

%Bibliography
\bibliographystyle{unsrt}  
\bibliography{references}

\begin{thebibliography}{10}

\bibitem{zghair_one_2021}
Noor Abdul~Khaleq Zghair and Ahmed~S. Al-Araji.
\newblock A one decade survey of autonomous mobile robot systems.
\newblock {\em International Journal of Electrical and Computer Engineering (IJECE)}, 11(6):4891, December 2021.

\bibitem{zhang_survey_2022}
Tianyao Zhang, Xiaoguang Hu, Jin Xiao, and Guofeng Zhang.
\newblock A survey of visual navigation: {From} geometry to embodied {AI}.
\newblock {\em Engineering Applications of Artificial Intelligence}, 114:105036, September 2022.

\bibitem{crespo_semantic_2020}
Jonathan Crespo, Jose~Carlos Castillo, Oscar~Martinez Mozos, and Ramon Barber.
\newblock Semantic {Information} for {Robot} {Navigation}: {A} {Survey}.
\newblock {\em Applied Sciences}, 10(2):497, January 2020.
\newblock Number: 2 Publisher: Multidisciplinary Digital Publishing Institute.

\bibitem{liu_comparative_2011}
Xiang Liu and Daoxiong Gong.
\newblock A comparative study of {A}-star algorithms for search and rescue in perfect maze.
\newblock In {\em 2011 {International} {Conference} on {Electric} {Information} and {Control} {Engineering}}, pages 24--27, April 2011.

\bibitem{soltani_path_2002}
A.~R. Soltani, H.~Tawfik, J.~Y. Goulermas, and T.~Fernando.
\newblock Path planning in construction sites: performance evaluation of the {Dijkstra}, {A}*, and {GA} search algorithms.
\newblock {\em Advanced Engineering Informatics}, 16(4):291--303, October 2002.

\bibitem{levine_learning_2023}
Sergey Levine and Dhruv Shah.
\newblock Learning robotic navigation from experience: principles, methods and recent results.
\newblock {\em Philosophical Transactions of the Royal Society B: Biological Sciences}, 378(1869):20210447, January 2023.

\bibitem{seymour_maast_2021}
Zachary Seymour, Kowshik Thopalli, Niluthpol Mithun, Han-Pang Chiu, Supun Samarasekera, and Rakesh Kumar.
\newblock {MaAST}: {Map} {Attention} with {Semantic} {Transformersfor} {Efficient} {Visual} {Navigation}, March 2021.
\newblock arXiv:2103.11374 [cs].

\bibitem{bowman_probabilistic_2017}
Sean~L. Bowman, Nikolay Atanasov, Kostas Daniilidis, and George~J. Pappas.
\newblock Probabilistic data association for semantic {SLAM}.
\newblock In {\em 2017 {IEEE} {International} {Conference} on {Robotics} and {Automation} ({ICRA})}, pages 1722--1729, May 2017.

\bibitem{drouilly_semantic_2015}
Romain Drouilly, Patrick Rives, and Benoit Morisset.
\newblock Semantic representation for navigation in large-scale environments.
\newblock In {\em 2015 {IEEE} {International} {Conference} on {Robotics} and {Automation} ({ICRA})}, pages 1106--1111, Seattle, WA, USA, May 2015. IEEE.

\bibitem{hermann_grounded_2017}
Karl~Moritz Hermann, Felix Hill, Simon Green, Fumin Wang, Ryan Faulkner, Hubert Soyer, David Szepesvari, Wojciech~Marian Czarnecki, Max Jaderberg, Denis Teplyashin, Marcus Wainwright, Chris Apps, Demis Hassabis, and Phil Blunsom.
\newblock Grounded {Language} {Learning} in a {Simulated} {3D} {World}, June 2017.
\newblock arXiv:1706.06551 [cs, stat].

\bibitem{ghaffari_grounding_2023}
Sadaf Ghaffari and Nikhil Krishnaswamy.
\newblock Grounding and {Distinguishing} {Conceptual} {Vocabulary} {Through} {Similarity} {Learning} in {Embodied} {Simulations}, May 2023.
\newblock arXiv:2305.13668 [cs].

\bibitem{lynch_interactive_2022}
Corey Lynch, Ayzaan Wahid, Jonathan Tompson, Tianli Ding, James Betker, Robert Baruch, Travis Armstrong, and Pete Florence.
\newblock Interactive {Language}: {Talking} to {Robots} in {Real} {Time}, October 2022.
\newblock arXiv:2210.06407 [cs].

\bibitem{bing_meta-reinforcement_2022}
Zhenshan Bing, Alexander Koch, Xiangtong Yao, Kai Huang, and Alois Knoll.
\newblock Meta-{Reinforcement} {Learning} via {Language} {Instructions}, September 2022.
\newblock arXiv:2209.04924 [cs].

\bibitem{fu_language_2019}
Justin Fu, Anoop Korattikara, Sergey Levine, and Sergio Guadarrama.
\newblock From {Language} to {Goals}: {Inverse} {Reinforcement} {Learning} for {Vision}-{Based} {Instruction} {Following}, February 2019.
\newblock arXiv:1902.07742 [cs, stat].

\bibitem{walter_learning_2013}
Matt Walter, Sachithra Hemachandra, Bianca Homberg, Stefanie Tellex, and Seth Teller.
\newblock Learning {Semantic} {Maps} from {Natural} {Language} {Descriptions}.
\newblock In {\em Robotics: {Science} and {Systems} {IX}}. Robotics: Science and Systems Foundation, June 2013.

\bibitem{gervet_navigating_2022}
Theophile Gervet, Soumith Chintala, Dhruv Batra, Jitendra Malik, and Devendra~Singh Chaplot.
\newblock Navigating to {Objects} in the {Real} {World}, December 2022.
\newblock arXiv:2212.00922 [cs].

\bibitem{studer_knowledge_1998}
Rudi Studer, V.~Richard Benjamins, and Dieter Fensel.
\newblock Knowledge engineering: principles and methods. {Data} {Knowl} {Eng} 25(1-2):161-197.
\newblock {\em Data \& Knowledge Engineering}, 25:161--197, March 1998.

\bibitem{guarino_formal_1998}
Nicola Guarino.
\newblock {\em Formal {Ontology} in {Information} {Systems}: {Proceedings} of the {First} {International} {Conference} ({FOIS}'98), {June} 6-8, {Trento}, {Italy}}.
\newblock IOS Press, 1998.
\newblock Google-Books-ID: Wf5p3\_fUxacC.

\bibitem{hemachandra_learning_2015}
Sachithra~Madhawa Hemachandra.
\newblock {\em Learning semantic maps from natural language}.
\newblock Thesis, Massachusetts Institute of Technology, 2015.
\newblock Accepted: 2015-07-17T19:12:02Z.

\bibitem{kostavelis_learning_2013}
Ioannis Kostavelis and Antonios Gasteratos.
\newblock Learning spatially semantic representations for cognitive robot navigation.
\newblock {\em Robotics and Autonomous Systems}, 61(12):1460--1475, December 2013.

\bibitem{han_semantic_2021}
Xiaoning Han, Shuailong Li, Xiaohui Wang, and Weijia Zhou.
\newblock Semantic {Mapping} for {Mobile} {Robots} in {Indoor} {Scenes}: {A} {Survey}.
\newblock {\em Information}, 12(2):92, February 2021.
\newblock Number: 2 Publisher: Multidisciplinary Digital Publishing Institute.

\bibitem{alenzi_semantic_2022}
Ziyad Alenzi, Emad Alenzi, Mohammad Alqasir, Majed Alruwaili, Tareq Alhmiedat, and Osama~Moh’d Alia.
\newblock A {Semantic} {Classification} {Approach} for {Indoor} {Robot} {Navigation}.
\newblock {\em Electronics}, 11(13):2063, January 2022.
\newblock Number: 13 Publisher: Multidisciplinary Digital Publishing Institute.

\bibitem{karimi_semantic_2021}
Sina Karimi, Rafael~Gomes Braga, Ivanka Iordanova, and David St-Onge.
\newblock Semantic {Navigation} {Using} {Building} {Information} on {Construction} {Sites}, April 2021.
\newblock arXiv:2104.10296 [cs].

\bibitem{jamieson_instance_2018}
Randall~K. Jamieson, Johnathan~E. Avery, Brendan~T. Johns, and Michael~N. Jones.
\newblock An {Instance} {Theory} of {Semantic} {Memory}.
\newblock {\em Computational Brain \& Behavior}, 1(2):119--136, June 2018.

\bibitem{louwerse_symbol_2011}
Max~M. Louwerse.
\newblock Symbol {Interdependency} in {Symbolic} and {Embodied} {Cognition}.
\newblock {\em Topics in Cognitive Science}, 3(2):273--302, April 2011.

\bibitem{guo_improve_2023}
Zhourui Guo, Meng Yao, Yang Yu, and Qiyue Yin.
\newblock Improve the efficiency of deep reinforcement learning through semantic exploration guided by natural language, September 2023.
\newblock arXiv:2309.11753 [cs].

\bibitem{nair_learning_2022}
Suraj Nair, Eric Mitchell, Kevin Chen, Brian Ichter, Silvio Savarese, and Chelsea Finn.
\newblock Learning {Language}-{Conditioned} {Robot} {Behavior} from {Offline} {Data} and {Crowd}-{Sourced} {Annotation}.
\newblock In {\em Proceedings of the 5th {Conference} on {Robot} {Learning}}, pages 1303--1315. PMLR, January 2022.
\newblock ISSN: 2640-3498.

\bibitem{zhou_language-conditioned_2024}
Hongkuan Zhou, Zhenshan Bing, Xiangtong Yao, Xiaojie Su, Chenguang Yang, Kai Huang, and Alois Knoll.
\newblock Language-{Conditioned} {Imitation} {Learning} with {Base} {Skill} {Priors} under {Unstructured} {Data}, February 2024.
\newblock arXiv:2305.19075 [cs].

\bibitem{goyal_pixl2r_2021}
Prasoon Goyal, Scott Niekum, and Raymond Mooney.
\newblock {PixL2R}: {Guiding} {Reinforcement} {Learning} {Using} {Natural} {Language} by {Mapping} {Pixels} to {Rewards}.
\newblock In {\em Proceedings of the 2020 {Conference} on {Robot} {Learning}}, pages 485--497. PMLR, October 2021.
\newblock ISSN: 2640-3498.

\bibitem{chan_zipfian_2022}
Stephanie C.~Y. Chan, Andrew~Kyle Lampinen, Pierre~Harvey Richemond, and Felix Hill.
\newblock Zipfian {Environments} for {Reinforcement} {Learning}.
\newblock In {\em Proceedings of {The} 1st {Conference} on {Lifelong} {Learning} {Agents}}, pages 406--429. PMLR, November 2022.
\newblock ISSN: 2640-3498.

\bibitem{louwerse_language_2009}
Max~M. Louwerse and Rolf~A. Zwaan.
\newblock Language {Encodes} {Geographical} {Information}.
\newblock {\em Cognitive Science}, 33(1):51--73, 2009.
\newblock \_eprint: https://onlinelibrary.wiley.com/doi/pdf/10.1111/j.1551-6709.2008.01003.x.

\bibitem{louwerse_representing_2012}
Max~M. Louwerse and Nick Benesh.
\newblock Representing {Spatial} {Structure} {Through} {Maps} and {Language}: {Lord} of the {Rings} {Encodes} the {Spatial} {Structure} of {Middle} {Earth}.
\newblock {\em Cognitive Science}, 36(8):1556--1569, 2012.
\newblock \_eprint: https://onlinelibrary.wiley.com/doi/pdf/10.1111/cogs.12000.

\bibitem{avery_reconstructing_2021}
Johnathan~E. Avery, Robert~L. Goldstone, and Michael~N. Jones.
\newblock Reconstructing maps from text.
\newblock {\em Cognitive Systems Research}, 70:101--108, December 2021.

\bibitem{avery_representation_2023}
Johnathan~E. Avery.
\newblock {\em Representation and {Retrieval} in {Semantic} {Memory}}.
\newblock PhD thesis, Indiana University-Bloomington, 2023.

\bibitem{de_vega_symbols_2012}
Manuel De~Vega, Arthur Glenberg, and Arthur Graesser.
\newblock {\em Symbols and embodiment: {Debates} on meaning and cognition}.
\newblock Oxford University Press, March 2012.

\bibitem{harnad_symbol_1990}
Stevan Harnad.
\newblock The {Symbol} {Grounding} {Problem}.
\newblock {\em Physica D: Nonlinear Phenomena}, 42(1-3):335--346, 1990.

\bibitem{pulvermuller_brain_2005}
Friedemann Pulvermüller, Yury Shtyrov, and Risto Ilmoniemi.
\newblock Brain {Signatures} of {Meaning} {Access} in {Action} {Word} {Recognition}.
\newblock {\em Journal of cognitive neuroscience}, 17:884--92, July 2005.

\bibitem{luo_transformer-based_2024}
Haonan Luo, Ziyu Guo, Zhenyu Wu, Fei Teng, and Tianrui Li.
\newblock Transformer-based vision-language alignment for robot navigation and question answering.
\newblock {\em Information Fusion}, 108:102351, August 2024.

\bibitem{huang_visual_2023}
Chenguang Huang, Oier Mees, Andy Zeng, and Wolfram Burgard.
\newblock Visual {Language} {Maps} for {Robot} {Navigation}, March 2023.
\newblock arXiv:2210.05714 [cs].

\end{thebibliography}

\end{document}